%% file: main.tex
\documentclass{article}
\usepackage{arydshln}
\usepackage[hidelinks]{hyperref}
\usepackage{spconf,amsmath,graphicx}
\usepackage[table]{xcolor}
\usepackage{graphicx}
\usepackage{amsmath}
\usepackage{amssymb}
\usepackage{enumitem}
\setlist{nosep, leftmargin=14pt}
\usepackage{subcaption}
\usepackage{booktabs}
\usepackage{lipsum}
\usepackage{mwe} 

\title{Fidelity-Imposed Displacement Editing for the Learn2Reg 2024 SHG-BF Challenge}

\name{
\begin{tabular}{c}
Jiacheng Wang$^{\star}$, Xiang Chen$^{\dagger}$, Renjiu Hu$^{\ddagger}$, Rongguang Wang$^{\mathsection}$, \\ 
Jiazheng Wang$^{\dagger}$, Min Liu$^{\dagger}$, Yaonan Wang$^{\dagger}$, Hang Zhang$^{\ddagger *}$
\end{tabular}
}

\address{$^{\star}$ Vanderbilt University,
    $^{\dagger}$ Hunan University
    $^{\ddagger}$ Cornell University
    $^{\mathsection}$ University of Pennsylvania
}

\begin{document}
%
\maketitle
\begingroup\renewcommand\thefootnote{\textasteriskcentered}
\footnotetext{Corresponding author. hz459@cornell.edu}
\endgroup

\begin{abstract}
Co-examination of second-harmonic generation (SHG) and bright-field (BF) microscopy enables the differentiation of tissue components and collagen fibers, aiding the analysis of human breast and pancreatic cancer tissues.
However, large discrepancies between SHG and BF images pose challenges for current learning-based registration models in aligning SHG to BF. 
In this paper, we propose a novel multi-modal registration framework that employs fidelity-imposed displacement editing to address these challenges.
The framework integrates batch-wise contrastive learning, feature-based pre-alignment, and instance-level optimization. 
Experimental results from the Learn2Reg COMULISglobe SHG-BF Challenge validate the effectiveness of our method, securing the 1st place on the online leaderboard. Our code is available at \url{https://github.com/JackyWang22/Neon_ShgBf}.

\end{abstract}

\begin{keywords}
Second-harmonic generation, Image registration, Contrastive learning.
\end{keywords}

\section{Introduction}
Image registration is a fundamental task in medical imaging, crucial for aligning images from different modalities or time points. 
Second-harmonic generation (SHG) microscopy provides high-resolution images sensitive to collagen fibers, while bright-field (BF) microscopy with hematoxylin and eosin (H\&E) staining highlights various tissue components \cite{eliceiri2021multimodal,keikhosravi2019intensity}. Accurate registration of SHG and BF images is essential for comprehensive cancer tissue analysis, offering deeper insights into tissue structure and pathology.

SHG-BF registration presents two primary challenges.
First, the two modalities differ significantly: SHG images emphasize collagen fibers, while BF images highlight stained tissue components, resulting in large visual discrepancies.
Second, the sparse distribution of highlighted structures in SHG creates a severe foreground-background imbalance, making it difficult for learning-based methods, including convolutional neural networks \cite{zhang2024memwarp,chen2024spatially,balakrishnan2019voxelmorph,zhang2024slicer}, vision transformers \cite{mok2022affine,chen2022transmorph}, and keypoint-based methods \cite{evan2022keymorph}, to perform effectively.


To address these challenges, we propose a novel SHG-BF multimodal registration method with the following key contributions:
\begin{enumerate}
    \item \textbf{Batch-wise contrastive loss (B-NCE)}:
    We introduce a batch-wise noise contrastive estimation loss to effectively capture shared features between SHG and BF images.
    \item \textbf{Feature-based prealignment and instance optimization}:
    A prealignment step using descriptor matching is followed by instance-level optimization to refine the registration.
    \item \textbf{Flexible transition of similarity metrics}:
    We combine local normalized cross-correlation (LNCC) and cross mutual information function (CMIF) as similarity metrics, balancing global and local alignment.
\end{enumerate}
The novel contrastive learning loss addresses the modality gap, while the instance optimization overcomes the foreground-background imbalance. 
Quantitative and qualitative results demonstrate considerable improvements in registration accuracy and robustness, earning us 1st place on the online \href{https://learn2reg.grand-challenge.org/evaluation/l2r24-comulisshgbf/leaderboard/}{leaderboard} of the Learn2Reg COMULISglobe SHG-BF Challenge.

\begin{figure}[!t]
    \centering
        \centering
        \includegraphics[width=0.8\columnwidth]{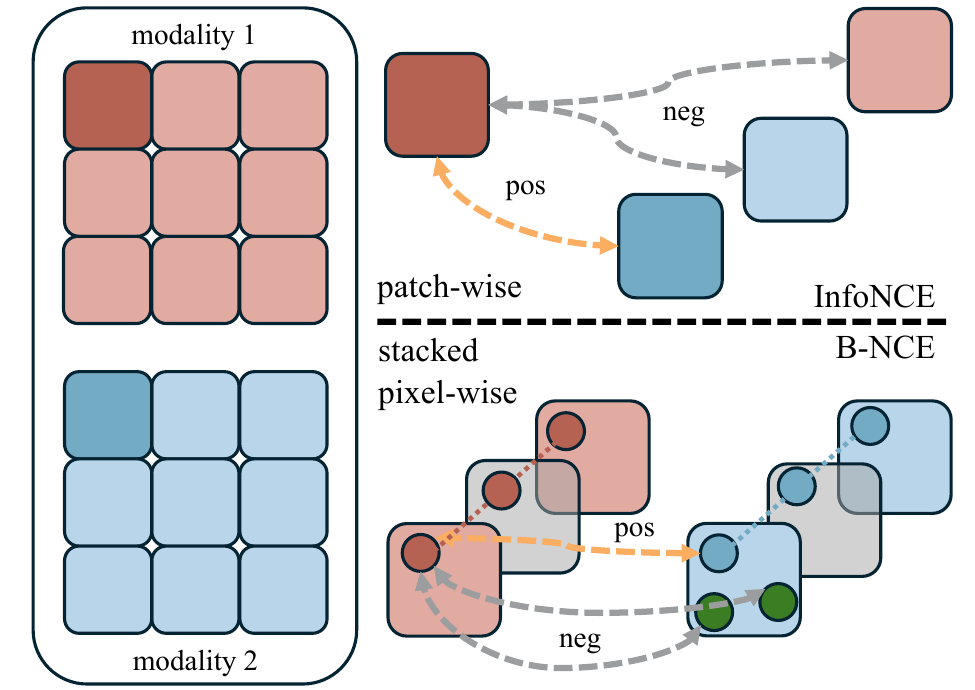}
        \caption{Illustration of our proposed batch-wise noise contrastive estimation (B-NCE) loss, which aggregates pixel-level information across patches, compared to traditional InfoNCE loss that operates directly on patch-level information.}
        \label{fig:bce_contrast}
    \end{figure}
\vspace{-9pt}

\section{Related Work}
\noindent\textbf{Classical methods.} Traditional registration methods typically formulate the problem as an iterative energy optimization, employing similarity criteria such as LNCC, mean-squared error (MSE), or mutual information (MI). 
Approaches like ANTS \cite{avants2009advanced} and ELASTIX \cite{klein2009elastix} utilize gradient descent for optimization but are prone to local minima, particularly when displacements exceed feature scales.
Discrete optimization techniques based on Markov Random Fields (MRF)~\cite{heinrich2013mrf} and local cost aggregation \cite{heinrich2015multi,steinbrucker2009large} have been proposed to mitigate these issues.

\noindent\textbf{Regression-Based Learning.}
Regression-based methods \cite{chen2021learning} train convolutional neural networks (ConvNets) to directly estimate the affine matrix. 
However, ConvNets lack inherent coordinate information, making affine matrix regression challenging. 
Vision transformers \cite{mok2022affine} and keypoint-based methods like KeyMorph \cite{evan2022keymorph} attempt to overcome these limitations by leveraging invariant feature representations, but still struggle with the high foreground-background imbalance.

\noindent\textbf{Descriptor Matching Learning.}
Descriptor matching methods \cite{detone2018superpoint,truong2019glampoints,wang2024novel,wang2024retinal} effectively address the high foreground-background imbalance by relying on feature descriptors and feature matching, utilizing both handcrafted and learning-based descriptors. 
Traditional RANSAC and modern graph- and attention-based approaches \cite{sarlin2020superglue} are commonly used for matching. 
However, these approaches lack a fidelity loss, leaving the output predictions unconstrained.

\begin{figure}[!t]
\centering
\includegraphics[width=\columnwidth]{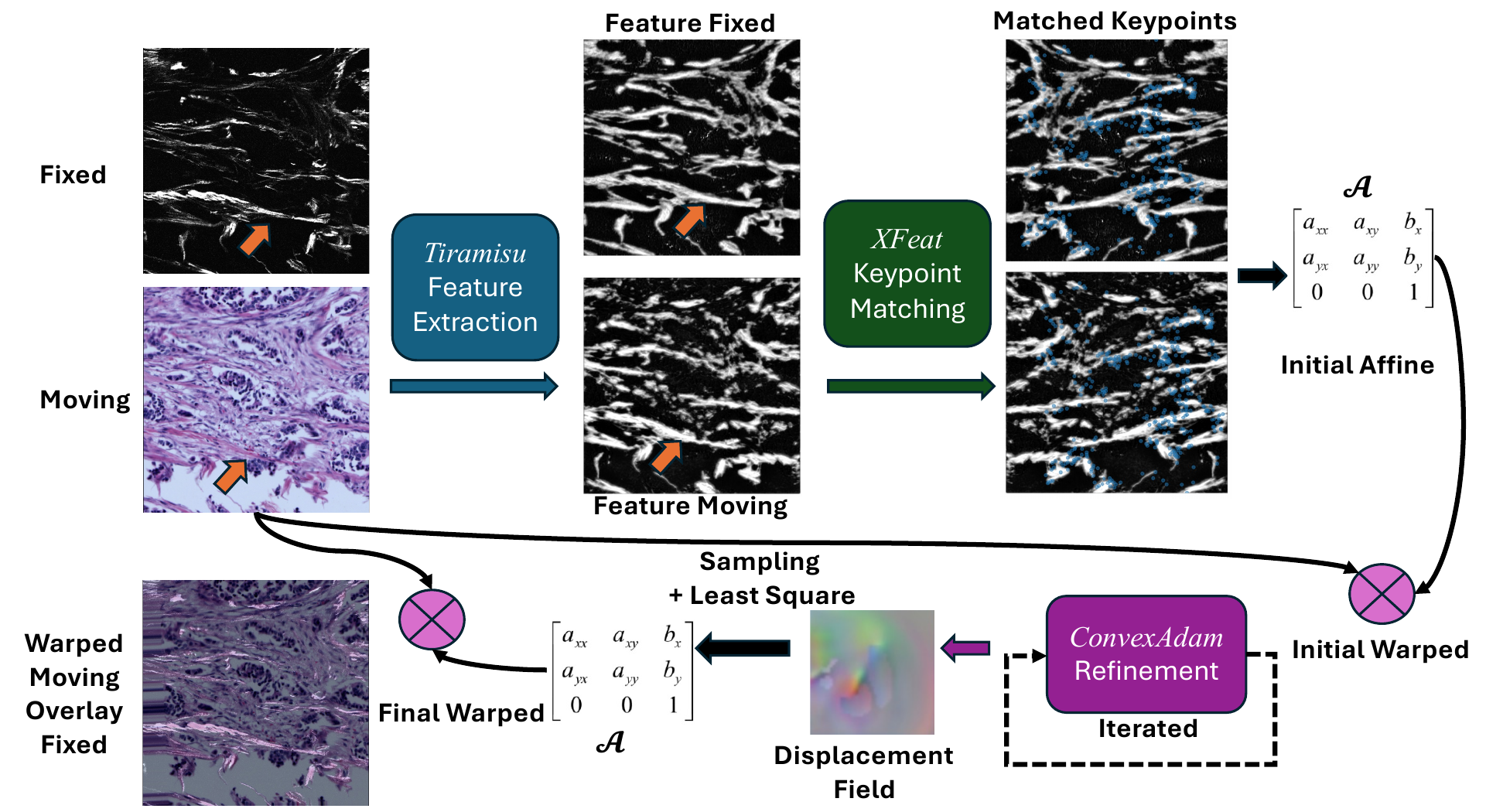}
\caption{Overview of our proposed framework, which follows three main steps: feature extraction, XFeat feature matching, and ConvexAdam fine-tuning.}
\centering
\label{fig:framework}
\end{figure}
\vspace{-2ex}

\section{Methods}
Our proposed registration framework, illustrated in Fig.~\ref{fig:framework}, consists of two main stages: a feature-based prealignment using contrastive representation learning and a test-time instance-level optimization for final registration. Initially, we learn a shared representation between SHG and BF images using our specially designed Batch-wise Noise Contrastive Estimation (B-NCE) loss. This shared latent space simplifies the multimodal registration problem into a monomodal one, making alignment more tractable. During inference, we refine the registration using ConvexAdam optimization, leveraging both cross-mutual information function (CMIF) and local normalized cross-correlation (LNCC) as similarity metrics to achieve accurate alignment.

Let \(I_A, I_B \in \mathbb{R}^2\) denote the fixed and moving images, respectively. Our goal is to construct affine transformations \(\Phi(I_A, I_B)\) for each image pair, mapping \(\mathbb{R}^2 \rightarrow \mathbb{R}^2\). We aim to align the transformed moving image to the fixed image, such that \(\Phi(I_A, I_B) \odot I_B \sim I_A\) where \(\odot\) denotes the spatial transformation of \(I_B\) using \(\Phi(I_A, I_B)\).


\subsection{Features prealignment with contrastive learning}

Due to the significant differences between SHG and BF images, direct registration is challenging. To address this, we map both images into a common latent space where shared features can be effectively captured. We achieve this using contrastive representation learning with a specially designed Batch-wise Noise Contrastive Estimation (B-NCE) loss.

\begin{figure}[!t]
    \centering
        \centering
        \includegraphics[width=\columnwidth]{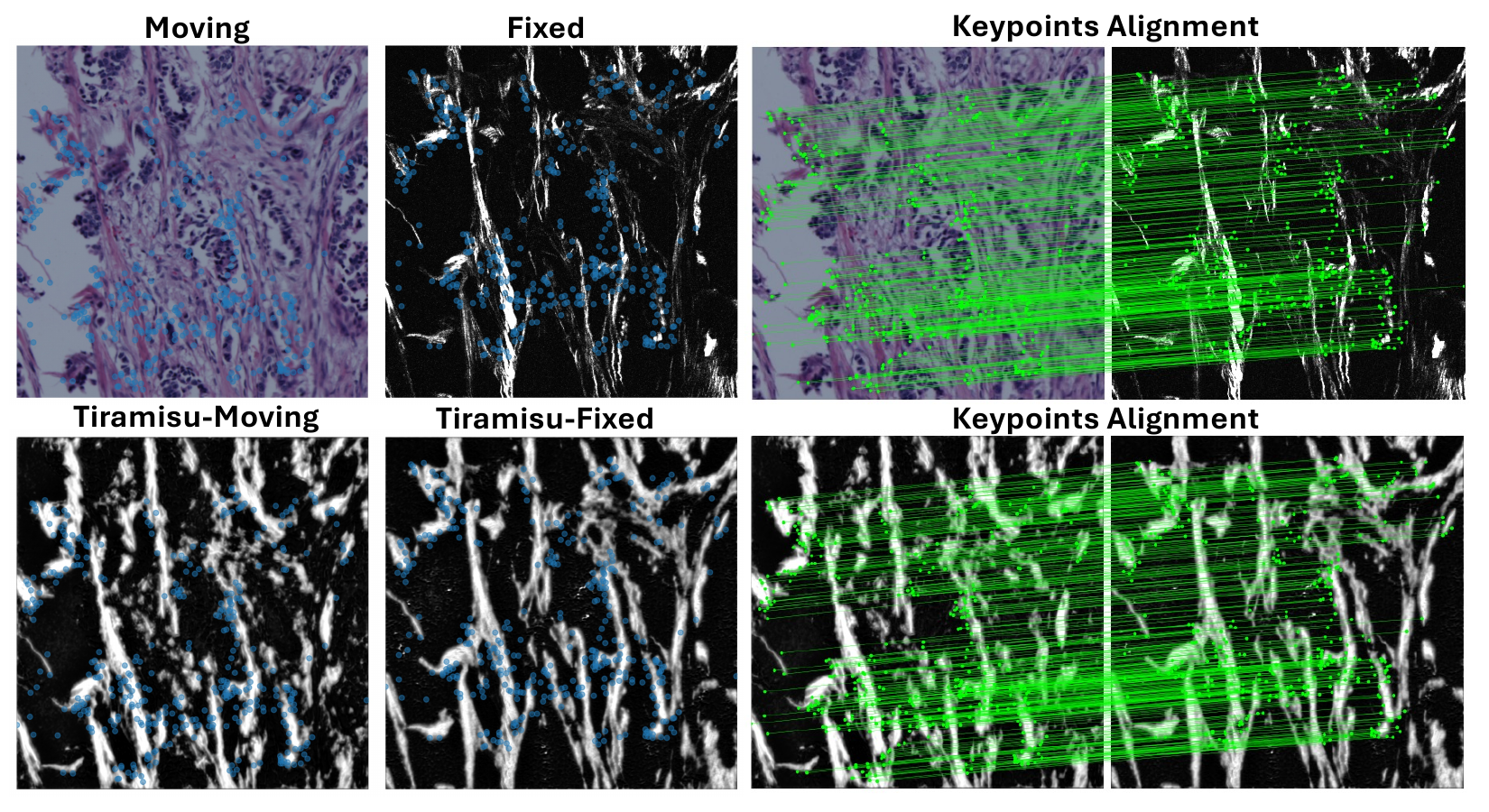}
        \caption{XFeat feature detection/description/matching in SHG/BF images and the Tiramisu feature images.}
        \label{fig:matching_keypoints}
    \end{figure}

Inspired by CoMIR \cite{pielawski2020comir}, we employ two identical Dense Tiramisu networks \cite{jegou2017one}, \(D_{\theta}\), to extract feature representations from \(I_A\) and \(I_B\):
$C_A = D_{\theta}(I_A)$ and $C_B = D_{\theta}(I_B)$,
where \(C_A, C_B \in \mathbb{R}^{H \times W \times C}\) and \(C\) is the number of feature channels. And we demonstrated the trained features shown in Fig.~\ref{fig:matching_keypoints} second row.

\noindent \textbf{Batch-wise noise contrastive estimation (B-NCE) loss.}
Our B-NCE loss is designed to enhance the network's ability to capture shared features between modalities by focusing on position-level similarities across the batch dimension. We extract patches from \(C_A\) and \(C_B\) of size \(p \times p\), resulting in sets of patches \(\{X_k\}\) and \(\{Y_k\}\) for \(k = 1, \dots, N\), where \(N\) is the number of patches.
To compute the loss, we consider the features at corresponding spatial locations across the batch:
\begin{equation}
\mathbf{x}_{i,j} = \{X_k(i,j)\}_{k=1}^{N}, \quad \mathbf{y}_{i,j} = \{Y_k(i,j)\}_{k=1}^{N},
\end{equation}
where \((i,j)\) indexes the spatial positions within each patch.

We define the similarity between features using the cosine similarity:
\begin{equation}
s(\mathbf{x}_{i,j}, \mathbf{y}_{i,j}) = \frac{\mathbf{x}_{i,j}^\top \mathbf{y}_{i,j}}{\|\mathbf{x}_{i,j}\| \|\mathbf{y}_{i,j}\|}.
\end{equation}

The B-NCE loss for each spatial position is:
\begin{equation}
\mathcal{L}_{\text{B-NCE}}^{i,j} = -\log \frac{\exp(s(\mathbf{x}_{i,j}, \mathbf{y}_{i,j}) / \tau)}{\sum_{k=1}^{N} \exp(s(\mathbf{x}_{i,j}, \mathbf{z}_k) / \tau)},
\end{equation}
where \(\mathbf{z}_k\) includes both positive and negative samples, and \(\tau\) is a temperature parameter.

The total loss is averaged over all spatial positions:
\begin{equation}
\mathcal{L}_{\text{B-NCE}} = \frac{1}{H_p W_p} \sum_{i=1}^{H_p} \sum_{j=1}^{W_p} \mathcal{L}_{\text{B-NCE}}^{i,j},
\end{equation}
where \(H_p\) and \(W_p\) are the height and width of the patches.

By minimizing \(\mathcal{L}_{\text{B-NCE}}\), the network learns to bring corresponding features from \(I_A\) and \(I_B\) closer in the latent space while pushing apart non-corresponding features, effectively capturing shared structures across modalities.

\noindent \textbf{Feature-based prealignment.}
With the learned representations \(C_A\) and \(C_B\), we perform feature-based prealignment. We detect keypoints and extract descriptors using a method inspired by XFeat~\cite{potje2024xfeat}, adapted to our context. We match features between \(C_A\) and \(C_B\) (Fig.~\ref{fig:matching_keypoints} second row) to compute an initial affine transformation \(\Phi_{\text{init}}\). We also demonstrated the matching keypoints in Fig.~\ref{fig:matching_keypoints} in blue dots and connected with green lines. 

\vspace{-5pt}

\vspace{-9pt} 

\subsection{Test-time Instance Optimization}

To address misalignment caused by local contrast noise in contrastive feature mapping, we adopt the core principles of the ConvexAdam method \cite{siebert2021fast}, enabling flexible instance optimization between affine and deformable transformations. Unlike the original ConvexAdam, which optimizes flow parameters using hand-crafted features, our approach leverages learned features from the preceding stage and employs a multi-resolution pyramid across scales of $1, \frac{1}{2}, \frac{1}{4}, \frac{1}{8}, \frac{1}{16}$. This hierarchical setup allows for efficient alignment at both large and fine scales.

Following the prealignment, we generate displacement fields using a convolutional network. Displacements are assigned by evaluating patch similarity costs and identifying minimal cost positions. Smooth regularization is then applied through Gaussian smoothing, as described in \cite{vercauteren2009diffeomorphic}. Our fidelity loss combines Local Normalized Cross-Correlation (LNCC), which handles local similarities and addresses the foreground-background imbalance, with Cross Mutual Information Function (CMIF), which considers global intensity relationships to bridge the modality gap.

Since the primary features in SHG images are sparsely distributed, we avoid using the deformation field directly. Instead, we sample displacements based on SHG intensity and perform least-squares analysis on these sparse displacements, resulting in an over-determined system of equations that we solve to obtain the final affine transformation. This approach allows for a smooth transition between deformable and affine transformations, enhancing the precision of the registration results.

\subsection{Multimodal Similarities: Mutual Information}
We compute mutual information between fixed image $I_A\subseteq X_A$ and moving image $I_B\subseteq X_B$ based on cross-mutual information function (CMIF)~\cite{pompe1998using, ofverstedt2022fast}.
Given images $I_A$ and $I_B$ intersecting on $I_{AB}$ and $I_{AB}\neq\emptyset$,
we firstly cluster each image $I_A$ and $I_B$ into level sets $\mathcal{A}\in\mathbb{Z}$ and $\mathcal{B}\in\mathbb{Z}$ accordingly using K-means.
For each level $a$ and $b$ from the level sets $\mathcal{A}$ and $\mathcal{B}$, we compute the marginal and joint histogram entries separately,
Then, based on the normalized histograms, we derive the marginal and joint Shannon entropies as follows,
\begin{equation}
H_A = -\sum_{a\in\mathcal{A}}\frac{I_A^a\cdot I_B}{N_{AB}}\log\frac{I_A^a\cdot I_B}{N_{AB}};
\end{equation}
\begin{equation}
H_B = -\sum_{b\in\mathcal{B}}\frac{I_A\cdot I_B^b}{N_{AB}}\log\frac{I_A\cdot I_B^b}{N_{AB}};
\end{equation}
\begin{equation}
H_{AB} = -\sum_{a\in\mathcal{A}}\sum_{b\in\mathcal{B}}\frac{I_A^a\cdot I_B^b}{N_{AB}}\log\frac{I_A^a\cdot I_B^b}{N_{AB}},
\end{equation}
where $I_A^a$ and $I_B^b$ represent images where the pixel values equal to level set value $a$ and $b$ respectively; $N_{AB}$ is the number of total pixels in both image $I_A$ and $I_B$.
Finally, we can compute the mutual information using $MI(I_A,I_B) = H_A + H_B-H_{AB}$.


\begin{figure}[!t]
\centering
\includegraphics[width=0.9\columnwidth]{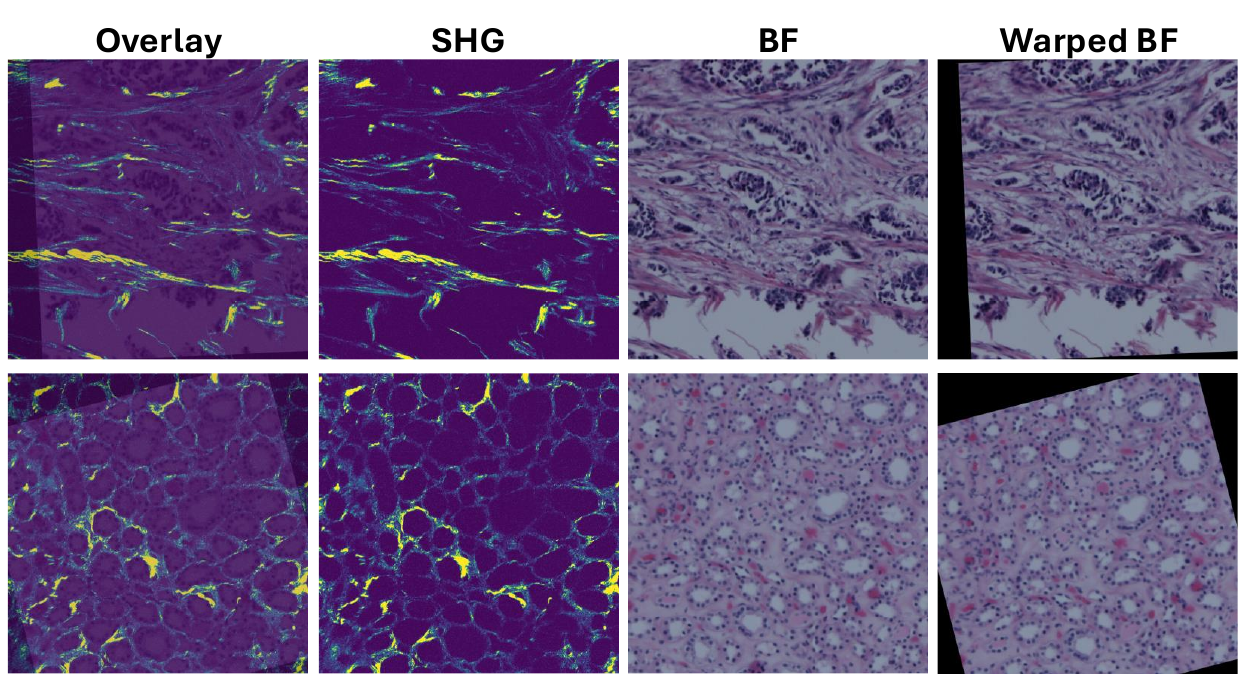}
\caption{Qualitative results on the validation set. From left to right: warped BF image overlaid on SHG image, SHG image (fixed) shown in Virdis colormap, BF image (moving), and warped BF image.} 
\label{fig:quali}
\end{figure} 
\vspace{-9pt}

\section{Experiments and Results}
\subsection{Dataset}
We evaluated our method on the dataset from the Learn2Reg 2024 Challenge Task 3: COMULISglobe SHG-BF~\cite{eliceiri2021multimodal}, which includes 156 training, 10 validation, and 40 test cases of SHG and H\&E-stained BF microscopy images of breast and pancreatic cancer tissues shown in Fig.~\ref{fig:quali}. Acquired at the University of Wisconsin-Madison, these images present significant challenges due to differences in imaging modalities: SHG images emphasize collagen fibers, while BF images highlight H\&E-stained tissue structures. Reliable expert annotations provide landmarks for validation and testing.


\subsection{Data Preprocessing and Evaluation Metrics}
For preprocessing, we applied stochastic intensity transformations using Bézier curves \cite{kobayashi2019bezier} and random affine augmentations, including scaling, rotation, and shearing, to enhance the robustness of our model to intensity variations and geometric distortions. During validation and testing, we omitted stochastic augmentations but maintained normalization and intensity adjustments to ensure consistency.

We used the Learn2Reg 2024 Challenge metrics, primarily Target Registration Error (TRE), to measure registration accuracy. TRE calculates the average distance between matched landmarks in the fixed and moving images—lower values mean better alignment. Validation results are from the challenge leaderboard.

\subsection{Quantitative Results}
Table~\ref{table:quant_shgbf} presents the performance of our method compared to top-performing methods from the challenge. Our approach achieved a mean TRE of 1.943 mm on the validation set, ranking first. Notably, our method demonstrated the lowest standard deviation (0.765 mm) among all submissions, indicating high robustness and consistency across different cases.

\input{docs/challenge_table}

To evaluate the effectiveness of our proposed components, we conducted ablation studies summarized in Table~\ref{table:quant_shgbf} (second part). LNCC is less sensitive to local noise and effectively handled the sparse SHG features, reducing the TRE by approximately 60\%. However, it is less capable of capturing global structure. In contrast, CMIF captures global structure better but is more sensitive to local noise introduced by the contrastive feature mapping, leading to worse results when used alone. By combining LNCC with CMIF in our fidelity loss, we balanced local robustness and global alignment, reducing the TRE from 2.361 to 1.943. This demonstrates that our fidelity-imposed displacement editing enhances registration accuracy by integrating the strengths of both metrics.

\begin{figure}[!t]
    \centering
        \centering
        \includegraphics[width=0.7\columnwidth]{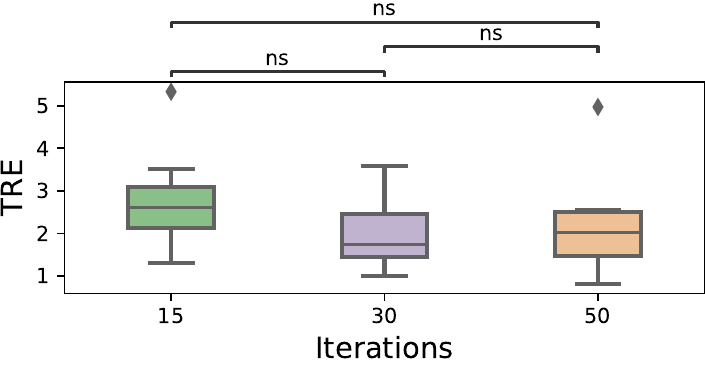}
        \caption{Impact of the number of optimization iterations on the mean TRE. Our method achieves consistent performance, with the best result at 30 iterations.}
        \label{fig:ablation}
    \end{figure}

Additionally, we tested different iteration counts during(15, 30, and 50 iterations) using both CMIF and LNCC (Fig.~\ref{fig:ablation}). Our method consistently achieved low TRE values, with the best performance at 30 iterations (mean TREs of 2.759, 1.943, and 2.168, respectively). This indicates our approach's stability without extensive computation.

\vspace{-11pt}

\subsection{Qualitative Results}
Two validation examples are presented in Fig.~\ref{fig:quali}. Despite significant structural differences due to the modality gap, our method successfully aligned the moving BF images to the fixed SHG images. The overlaid warped BF images show main structures aligned with minimal misalignment, demonstrating effective multimodal registration.

Figure~\ref{fig:matching_keypoints} shows feature matching with our learned representations. Despite large modality discrepancies, the features are visually similar, enabling robust matching. The sufficient matching pairs confirm the effectiveness of B-NCE loss in bridging the modality gap for feature-based prealignment.

\section{Conclusion}
In this paper, we propose a new affine registration framework that integrates a novel contrastive learning loss (B-NCE) with rich contextual information from modern neural networks inside fidelity-imposed iterative instance optimization. 
Our fidelity-imposed network editing framework overcomes modality gap between SHG and BF images, providing robust and accurate SHG-BF registration performance. 
Experimental results from the Learn2Reg COMULISglobe SHG-BF Challenge demonstrate the effectiveness of our approach, earning the 1st place on the online leaderboard.


\noindent\textbf{Acknowledgment.} This work is supported, in part, by the National Key Research and Development Program of China under Grant 2022YFE0134700, National Natural Science Foundation of China under Grant U22B2050 and 62425305, and the Science and Technology Program of Changsha under Grant kq2102009. \textbf{Compliance with Ethical Standards.}
Ethical approval was not required as confirmed by the license attached with the open access data. 

{
\bibliographystyle{IEEEbib}
\bibliography{ref.bib}
}

\end{document}

%% file: docs/challenge_table.tex
\begin{table}[!t]
    \centering
    \caption{TRE comparison of teams on the Learn2Reg 2024 Challenge validation leaderboard. 
    Rankings are based on a snapshot of the leaderboard as of Sep 20, 2024, excluding challenge organizer entries.}
    \label{table:quant_shgbf}
    \resizebox{0.95\columnwidth}{!}{
    \begin{tabular}{lcc}
        \hline
        \hline
        Method & TRE (LM) & Best Validation Place \\
        \hline
        Team VROC & 2.620 $\pm$ 1.206 & 36th \\
        Team Yangzhao & 2.578 $\pm$ 1.377 & 34th \\
        Team IWM & 2.077 $\pm$ 1.182 & 8th \\
        Team meeem & \underline{2.032} $\pm$ \underline{1.026} & 6th \\
        \hline
        XFeat Only & 5.939 $\pm$ 8.198 & 39th \\
        XFeat~+~Ours (CMIF) & 9.128 $\pm$ 1.211 & 72th\\
        XFeat~+~Ours (LNCC) & 2.361 $\pm$ 0.812 & 20th \\
        \rowcolor{lightgray}
        XFeat~+~Ours (LNCCC+CMIF) & \textbf{1.943} $\pm$ \textbf{0.765} & 1st \\
        \hline
        \hline
    \end{tabular}
    }
\end{table}

%% file: main.bbl
\begin{thebibliography}{10}

\bibitem{eliceiri2021multimodal}
Kevin Eliceiri et~al.,
\newblock ``Multimodal biomedical dataset for evaluating registration methods (full-size tma cores),''
\newblock {\em Zenodo, Feb}, 2021.

\bibitem{keikhosravi2019intensity}
Adib Keikhosravi et~al.,
\newblock ``Intensity-based registration of bright-field and second-harmonic generation images of histopathology tissue sections,''
\newblock {\em Biomedical Optics Express}, vol. 11, no. 1, pp. 160--173, 2019.

\bibitem{zhang2024memwarp}
Hang Zhang et~al.,
\newblock ``Memwarp: Discontinuity-preserving cardiac registration with memorized anatomical filters,''
\newblock in {\em MICCAI}. Springer, 2024, pp. 671--681.

\bibitem{chen2024spatially}
Xiang Chen et~al.,
\newblock ``Spatially covariant image registration with text prompts,''
\newblock {\em IEEE Transactions on Neural Networks and Learning Systems}, 2024.

\bibitem{balakrishnan2019voxelmorph}
Guha Balakrishnan et~al.,
\newblock ``Voxelmorph: a learning framework for deformable medical image registration,''
\newblock {\em IEEE TMI}, vol. 38, no. 8, pp. 1788--1800, 2019.

\bibitem{zhang2024slicer}
Hang Zhang et~al.,
\newblock ``Slicer networks,''
\newblock {\em arXiv preprint arXiv:2401.09833}, 2024.

\bibitem{mok2022affine}
Tony~CW Mok and Albert Chung,
\newblock ``Affine medical image registration with coarse-to-fine vision transformer,''
\newblock in {\em CVPR}, 2022, pp. 20835--20844.

\bibitem{chen2022transmorph}
Junyu Chen et~al.,
\newblock ``Transmorph: Transformer for unsupervised medical image registration,''
\newblock {\em Medical image analysis}, vol. 82, pp. 102615, 2022.

\bibitem{evan2022keymorph}
M~Yu Evan et~al.,
\newblock ``Keymorph: Robust multi-modal affine registration via unsupervised keypoint detection,''
\newblock in {\em MIDL}, 2022.

\bibitem{avants2009advanced}
B~B Avants et~al.,
\newblock ``Advanced normalization tools (ants),''
\newblock {\em Insight j}, vol. 2, no. 365, pp. 1--35, 2009.

\bibitem{klein2009elastix}
Stefan Klein et~al.,
\newblock ``Elastix: a toolbox for intensity-based medical image registration,''
\newblock {\em IEEE TMI}, vol. 29, no. 1, pp. 196--205, 2009.

\bibitem{heinrich2013mrf}
Mattias~P Heinrich et~al.,
\newblock ``Mrf-based deformable registration and ventilation estimation of lung ct,''
\newblock {\em IEEE TMI}, vol. 32, no. 7, pp. 1239--1248, 2013.

\bibitem{heinrich2015multi}
Mattias~P Heinrich et~al.,
\newblock ``Multi-modal multi-atlas segmentation using discrete optimisation and self-similarities.,''
\newblock {\em ISBI}, vol. 1390, pp. 27, 2015.

\bibitem{steinbrucker2009large}
Frank Steinbr{\"u}cker et~al.,
\newblock ``Large displacement optical flow computation without warping,''
\newblock in {\em ICCV}, 2009.

\bibitem{chen2021learning}
Xu~Chen et~al.,
\newblock ``Learning unsupervised parameter-specific affine transformation for medical images registration,''
\newblock in {\em MICCAI}. Springer, 2021, pp. 24--34.

\bibitem{detone2018superpoint}
Daniel DeTone et~al.,
\newblock ``Superpoint: Self-supervised interest point detection and description,''
\newblock in {\em CVPR workshops}, 2018, pp. 224--236.

\bibitem{truong2019glampoints}
Prune Truong et~al.,
\newblock ``Glampoints: Greedily learned accurate match points,''
\newblock in {\em ICCV}, 2019, pp. 10732--10741.

\bibitem{wang2024novel}
Jiacheng Wang et~al.,
\newblock ``Novel oct mosaicking pipeline with feature-and pixel-based registration,''
\newblock in {\em ISBI}. IEEE, 2024, pp. 1--5.

\bibitem{wang2024retinal}
Jiacheng Wang et~al.,
\newblock ``Retinal ipa: Iterative keypoints alignment for multimodal retinal imaging,''
\newblock {\em arXiv preprint arXiv:2407.18362}, 2024.

\bibitem{sarlin2020superglue}
Paul-Edouard Sarlin et~al.,
\newblock ``Superglue: Learning feature matching with graph neural networks,''
\newblock in {\em CVPR}, 2020, pp. 4938--4947.

\bibitem{pielawski2020comir}
Nicolas Pielawski et~al.,
\newblock ``Comir: Contrastive multimodal image representation for registration,''
\newblock {\em Advances in neural information processing systems}, vol. 33, pp. 18433--18444, 2020.

\bibitem{jegou2017one}
Simon J{\'e}gou et~al.,
\newblock ``The one hundred layers tiramisu: Fully convolutional densenets for semantic segmentation,''
\newblock in {\em CVPR workshops}, 2017, pp. 11--19.

\bibitem{potje2024xfeat}
Guilherme Potje et~al.,
\newblock ``Xfeat: Accelerated features for lightweight image matching,''
\newblock in {\em CVPR}, 2024.

\bibitem{siebert2021fast}
Hanna Siebert et~al.,
\newblock ``Fast 3d registration with accurate optimisation and little learning for learn2reg 2021,''
\newblock in {\em MICCAI}. Springer, 2021, pp. 174--179.

\bibitem{vercauteren2009diffeomorphic}
Tom Vercauteren et~al.,
\newblock ``Diffeomorphic demons: Efficient non-parametric image registration,''
\newblock {\em NeuroImage}, vol. 45, no. 1, pp. S61--S72, 2009.

\bibitem{pompe1998using}
Bernd Pompe et~al.,
\newblock ``Using mutual information to measure coupling in the cardiorespiratory system,''
\newblock {\em IEEE Engineering in Medicine and Biology Magazine}, vol. 17, no. 6, pp. 32--39, 1998.

\bibitem{ofverstedt2022fast}
Johan {\"O}fverstedt et~al.,
\newblock ``Fast computation of mutual information in the frequency domain with applications to global multimodal image alignment,''
\newblock {\em Pattern Recognition Letters}, vol. 159, pp. 196--203, 2022.

\bibitem{kobayashi2019bezier}
Ken Kobayashi et~al.,
\newblock ``B{\'e}zier simplex fitting: Describing pareto fronts of simplicial problems with small samples in multi-objective optimization,''
\newblock in {\em AAAI}, 2019, vol.~33, pp. 2304--2313.

\end{thebibliography}
